\documentclass{article}
\usepackage[utf8]{inputenc}
\usepackage{glossaries}
\usepackage[dvipsnames]{xcolor}
\usepackage{todonotes}
\usepackage{booktabs}
\usepackage{array}
\usepackage{amsmath}
\usepackage{graphicx}
\usepackage{authblk}
\usepackage{tikz}
\newacronym{wds}{WDS}{water distribution systems}
\newacronym{udf}{UDF}{unidirectional flushing}
\newacronym{cf}{CF}{conventional flushing}
\newacronym{afd}{AFD}{automatic flushing device}
\newacronym{spot}{SPOT}{Sequential Parameter Optimization Toolbox}
\newacronym{minlp}{MINLP}{mixed-integer nonlinear programming}
\title{Technical Report: Flushing Strategies in Drinking Water Systems}
\author{Margarita Rebolledo}
\author{Sowmya Chandrasekaran}
\author{Thomas Bartz-Beielstein}
\affil{Institute for Data Science, Engineering, and Analytics, TH K\"oln. Germany}
\date{December 2020}

\begin{document}

\maketitle

\section{Introduction}
Drinking water supply and distribution systems are critical infrastructure that has to be protected for the safety of the public health. 
Water systems generally include storage tanks, pipes, pumps, valves, reservoirs, meters, fittings, and other hydraulic appliances.
Considering the complexity and the widespread coverage it is very hard to maintain the distribution network to meet the desired water quality level. One important tool in the maintenance of \gls{wds} is flushing.

Flushing is a process carried out in a periodic fashion to clean sediments and other contaminants in the water pipes. Also, in-case of an accidental or intentional contamination event, an emergency flushing plan has to be executed. A systematic and well designed flushing strategy improves the quality of the drinking water and also helps in maintaining the complex and costly water infrastructure. A complete overview of water security systems, possible examples of intentional and unintentional contamination, and the importance of maintaining a good \gls{wds} is discussed extensively in \cite{Jank14a}.
A general guidance for a flushing process is given in \cite{Fried02a} where the authors present a four step flushing program that should be followed when executing a flushing on the network independently from the magnitude or frequency needed for a particular system. 

A flushing strategy can be formulated as a single-objective or multi-objective optimization problem based on its final purpose. Some of the possible goals include minimizing the water requirements for flushing, minimizing the impact on the public, minimizing the cost involved, minimizing the resource utilization, maximizing deposits detachment, optimizing the flushing path or optimizing the number of hydrants to be used. \\


In this report a non-exhaustive overview of optimization methods for flushing in \gls{wds} is given.  
The structure of the report is as follows:
The different flushing strategies are discussed in Section \ref{sec:2}. In Section \ref{sec:3}, various optimization methods for flushing found in the literature are described. The article concludes with short summary and discussion in Section \ref{sec:4}.

\section{Flushing Techniques}\label{sec:2}

Given the diversity in network's size, architecture or use intensity, there is no universal flushing program. When performing a flushing process
there are three main flushing techniques available \cite{Anto99a}:  conventional, unidirectional and continuous blow-off. The flushing goal determines which technique is best to use.\\

Traditionally, \Gls{cf} has been carried out. It consist of opening hydrants in specific locations to let the water flow out of the pipe. This action doesn't usually includes the activation of isolation valves, which are valves used to focus the water flow into only one pipeline path in order to increase flow velocity. Usually the intended effect of \gls{cf} is to replace the old water found in the pipes with fresher water. The criteria to stop the flow of water out of the hydrant tends to be a visual evaluation of water's turbidity  or quality measurements from sensors in the vicinity. \gls{cf} does not require much preparation or planing.

Later, \gls{udf} was developed \cite{Ober94a} where a section of the pipe system is isolated by closing key valves with the objective of creating a single directional flow with a velocity of at least 1.5 m/s. This minimal velocity is necessary to remove the impurities on the pipes walls and can vary according to pipe sizing and amount of impurities present. The implementation of \gls{udf} requires more extensive planing and preparation in defining the areas or pipes  in which it will be applied. This strategy is recommended as with higher velocity, better water quality can be achieved with less water utilization \cite{Poul10a}. 

Continuous blow-off \cite{Fried02a} is a technique used mainly on locations with low water circulation, especially at dead ends. Through the use of blow-off valves a portion of the water is extracted in a low velocity flow. This technique is not appropriate to remove sediments in the pipes but can be helpful in, for example, stabilizing the disinfectant levels by removing stagnate water.\\

\section{Planning and Optimization methods}\label{sec:3}

To implement a flushing strategy, a good planning and optimal utilization of time and resources are essential. In this section we present some of the studies and strategies found in the literature that focus first on the problem of planing a flushing and second on optimizing the flushing.\\

In \cite{Bara08a}, an optimization based strategy was proposed to optimally deal with drinking water contamination. 
As the resulting optimization problem was a \gls{minlp} problem, genetic algorithm was used to identify the optimal node, to alter the demand, to modify demands for the nodes, and, to identify optimal location of pipe closures. 
The authors proposed that the genetic algorithm can solve such a complicated problem and find optimal solutions to minimize the impact of a contamination event in a distribution network. 
The results from the optimization problem also allowed for optimal allocation of water resources in addition to the spread of the contamination. As the authors point out, the performance of the genetic algorithm is dependent on the tuning of the algorithmic parameters. 
Specialized parameter optimization techniques such as \gls{spot}\cite{Bart10a} can be used to resolve issues associated with manual tuning of the parameters. \\

In \cite{Leon10a}, optimization strategies using both single objectives and multiple objectives were investigated. Again, owing to the complexity of the resulting optimization problem, the authors employed genetic algorithm to obtain the optimal solutions. The authors considered two objectives: minimization to the damage to public health and minimization of operational costs. The authors employed the single objective formulation as the weighted multiple objectives. As an improvement to \cite{Bara08a}, the authors provided the strategies to obtain the flushing paths along with different flushing methods. The authors found that the computation times offered by genetic algorithms can be prohibitively large for very large networks and propose other random search algorithms. To compare algorithms, a novel comparison framework is proposed in \cite{Chand20a} that offers a stringent comparison between two stochastic optimization algorithms. The result from \cite{Chand20a} can be used by the practitioners to measure the performance improvement.

In \cite{Poul10a}, a heuristic set of rules were proposed for \gls{udf} in the event of water contamination. The flushing procedure is initiated once the water contamination is identified and the source of the contamination is eliminated. Before proceeding with the flushing, the contaminated area is isolated and public advisories to the affected parties are transmitted. Depending on the type of the contamination, either a non-adherent to piping infrastructure flushing or adherent to piping infrastructure flushing is initiated. Once the flushing strategy is initiated, it is continuously monitored to eliminate all contaminants in the affected pipelines. The flushing strategy is applied with promising results to two experimental networks. 

\cite{Deue14a} extends the approach proposed in \cite{Leon10a} by offering more flexibility in the flushing of sub-networks and proposes a tool for planning and optimization of \gls{udf} called \emph{Flushing Planner}. An optimal plan was proposed considering the effort required for valve manipulations and thereby minimizing the efforts of operational staffs.

\cite{Bara19a} proposes a strategy similar to \cite{Poul10a} in terms of the sequence of the flushing strategy: detect the contamination, public notification, and isolation and containment. In addition, \cite{Bara19a}  also proposes an optimization method to select the best response among the possible actions proposed in \cite{Bara08a}. \\

Flushing strategies for optimization during the flushing process can also be found in the bibliography.
In \cite{Fove12a} the focus is on open-channel networks and the problem of algae grow which can clog hydraulic devices and cause potential sanitary risks. The algae can be cleaned from the network by performing regular flushing but the amount of detached algae should remain under a certain threshold in order to avoid clogging. In order to control the amount of detached algae the water turbidity can be used as an indirect measure. 
The proposed flushing strategy is based on a quasi-linear model for the algae detachment that is designed to maximize the amount of detached algae while complying with two constraint: maintain the turbidity under a given threshold and minimize the volume of water required.

In \cite{Xie15a} the optimization of the flushing operation needed to maintain chlorine levels is discussed. The proposed method utilizes \gls{afd} to control pipe flushing via \gls{cf}.
One factor needed to maintain water quality in \gls{wds} is to control and maintain a sufficient chlorine residual to act as a disinfectant.
Flushing the pipes in order to remove stagnate water is one of the available methods to maintain a high enough chlorine residual level. 
The aim of the proposed strategy is to diminish the flushed water volume by optimizing the activation times of \glspl{afd} in a network. The method is divided in three phases: in the first phase the \glspl{afd} present in the network are tested via a hydraulics model to test if the desired chlorine level can be maintained by their activation/deactivation. In the second phase the hourly activation/deactivation of the individual \gls{afd} reformulated as a constantly activated \gls{afd} with a lower flow capacity in order to reduce the problem solution's space. Finally in the third phase simulated annealing is implemented to find the optimal solution on the solution space. The three-phase method was simulated using a real network topology and results indicated that the optimized results used less water volume that the current manual practice.

In \cite{Tang17a} the authors focus on 
the loss in water quality originated by the accumulation of residuals due to sedimentation on the pipe wall also named growth-ring. The removal of the growth-ring via two-phase flushing is studied using a small experimental setup and comparing its performance in terms of water savings to single-phase flushing. The best experimental result setups (air inflow/cutoff frequency and inlet nozzle shape) are tested on a real pipeline. The results show that two-phase flushing present significant saving in terms of water volume and flushing time.

\section{Discussion and Summary}\label{sec:4}

In this article, various flushing strategies with single or multiple objectives are studied. 
Given the vastly different topographies and utilization frequencies of \gls{wds} no single optimization configuration for flushing planing can be implemented.
A common issue in the optimization strategies is the problem of manual tuning of the parameters of the optimization algorithm. We suggest automatic parameter tuning toolbox \gls{spot} as a possible solution to overcome this issue \cite{Bart10a}. Also, in-order to choose the most appropriate algorithm for flushing, we suggest the use of statistical framework algCompare \cite{Chand20a}. This framework stringently evaluates the performance improvement achieved considering two main criteria: statistical significance and practical relevance.  Based on the desired/ meaningful performance requirement for this specific application of flushing, the performances of various flushing optimization algorithms can be evaluated. \\

Water quality in \gls{wds} plays a decisive factor in determining the appropriate frequency of maintenance flushing. 
An article published by \cite{Carr05a} studies the composition of the sediments removed during \gls{udf} in four Canadian Networks and proposes a flushing frequency based on the amount of deposits measured. The authors found a correlation between sediment accumulation and water turbidity, and recommend a continuous monitoring of turbidity coupled with a mineral analysis. Such analysis and monitoring can help assess the cost-benefit of \gls{udf} and possibly influence the frequency of the flushing to better adapt to the needs of the network.

Sediment accumulation was also studied in \cite{Blok15a}. The authors studied particle accumulation rate in relation to incoming water quality in two similar \gls{wds}. The basis of this study is given by \cite{Moun14a} where a data-driven model is used to estimate the rate of sediment material accumulation by investigating its correlation with certain network characteristics like water properties and pipe material. In \cite{Blok15a} it is demonstrated that, for otherwise similar networks, the rate of sediment accumulation is also heavily influenced by the quality of the water coming into the network.

This makes it clear that the proposal of an optimizer for the flushing process needs to have a clear definition of the objective that the flushing needs to achieve, for example maintain chlorine levels or a certain level of turbidity, and the exact characteristics of the network where it should be implemented.


\section*{Acknowledgements}
This research work is funded by project \textit{Open Water Open Source(OWOS)} (reference number: 005-1703-0011) and is kindly supported by the FH Zeit f{\"u}r  Forschung, Ministerium fur Innovation, Wissenschaft und Forschung des Landes NRW.
 \\
\bibliographystyle{unsrt}
\bibliography{flushBib}

\end{document}